\documentclass[11pt]{article}

\usepackage[hyperref]{EACL2023}

\usepackage{times}
\usepackage{latexsym}
\usepackage{amsfonts}
\usepackage{amsmath}
\usepackage{dsfont}
\usepackage{bm}
\usepackage{graphicx}
\usepackage{dirtytalk}
\usepackage{microtype}
\usepackage{booktabs}
\usepackage{algorithm}
\usepackage{algpseudocode}
\usepackage{caption}
\usepackage{subcaption}

\graphicspath{ {./images/} }

\renewcommand{\vec}[1]{\bm{#1}}

\newcommand{\vE}{\ensuremath \vec{E}}

\DeclareMathOperator*{\tok}{tok}

\DeclareMathOperator*{\argmin}{arg\,min}

\newcommand{\sage}[0]{\textsc{SaGe}}
\newcommand{\bpe}[0]{\textsc{BPE}}
\newcommand{\unilm}[0]{\textsc{UnigramLM}}
\newcommand{\unigramlm}[0]{\textsc{UnigramLM}}

\newcommand{\skipgram}[0]{\textsc{SkipGram}}

\newcommand{\todo}[1]{}

\title{Incorporating Context into Subword Vocabularies}

\author{Shaked Yehezkel \\
  Blavatnik School of Computer Science \\
  Tel-Aviv University \\
  Tel-Aviv, Israel \\
\texttt{shakedy@mail.tau.ac.il} \\\And
  Yuval Pinter \\
  Department of Computer Science \\
  Ben-Gurion University of the Negev \\
  Beer Sheva, Israel \\
\texttt{uvp@cs.bgu.ac.il} \\}

\begin{document}

\maketitle

\begin{abstract}
  Most current popular subword tokenizers are trained based on word frequency statistics over a corpus, without considering information about co-occurrence or context.
  Nevertheless, the resulting vocabularies are used in language models' highly contextualized settings.
  We present \sage{}, a tokenizer that tailors subwords for their downstream use by baking in the contextualized signal at the vocabulary creation phase.
  We show that \sage{} does a better job than current widespread tokenizers in keeping token contexts cohesive, while not incurring a large price in terms of encoding efficiency or domain robustness.
  \sage{} improves performance on English GLUE classification tasks as well as on NER, and on Inference and NER in Turkish, demonstrating its robustness to language properties such as morphological exponence and agglutination.
\end{abstract}

\section{Introduction}
\label{sec:intro}

Much of the research space in current NLP is focused on advancing models: modifying pre-training objectives, improving network architectures, adding tasks and schemes for downstream evaluation.
Limited work is dedicated to a crucial step underlying all modern large language models (LLMs), namely the \textbf{tokenization} phase.
In order to process a given string of text, an LLM must first obtain a vector representation of the input by segmenting it into tokens.
Since out-of-vocabulary (OOV) items inhibit the performance of models, current tokenizers produce tokens which are possibly proper subsegments of input words, known as \textbf{subwords}.
This method, popularized by systems such as WordPiece~\cite{schuster2012japanese}, Byte-Pair Encoding~\cite[\bpe;][]{sennrich-etal-2016-neural} and \unigramlm~\cite{kudo-2018-subword}, allows any word to be represented by one or more tokens, removing the OOV problem while allowing more flexibility in determining the token vocabulary size, which ultimately affects model speed (mostly through the softmax generation targets) and performance (through better ability to represent less-frequent words).

\begin{table}
    \centering
    \footnotesize
    \begin{tabular}{lp{6cm}}
        \toprule
        \bpe{} & \_His \_son \_Raj ash ri \_Sud h ak ar \_has \_p enn ed \textbf{\_dial og ues} \_and \_songs \_for \_some \_films \_that \_were \_dubbed \_into \_Telugu . \\
        \sage{} & \_His \_son \_Raj ash r i \_Sud h a k a r \_has \_penn e d \textbf{\_dial ogues} \_and \_songs \_for \_some \_films \_that \_were \_dubbed \_into \_Telugu . \\ \midrule
        \bpe{} & \_This \_gene \_is \_a \textbf{\_pseud og ene} \_in \_humans \_and \_most \_other \_prim ates . \\
        \sage{} & \_This \_gene \_is \_a \textbf{\_pseud ogene} \_in \_humans \_and \_most \_other \_prim ates . \\ \midrule
        \bpe{} & \_The \textbf{\_St o og es} \_work \_for \_Mir acle \_Det ective \_Agency , \\
        \sage{} & \_The \textbf{\_St o o g e s} \_work \_for \_Mir acle \_Det ective \_Agency , \\
        \bottomrule
    \end{tabular}
    \caption{The token \emph{og} is selected by \bpe{} (vocabulary of size 16,000) for achieving the frequency objective, but is discarded by \sage{} for failing to be contextually coherent. These examples from the corpus demonstrate some different contexts.}
    \label{tab:examples}
\end{table}

One potential pitfall of both \bpe{} and \unigramlm, as well as their proposed variants~\cite{he-etal-2020-dynamic,provilkov-etal-2020-bpe}, is that they are trained on word frequency statistics alone, without considering information about word co-occurrence or contexts.
At the same time, the resulting vocabularies are used in highly contextualized settings, the LLMs, where a single subword such as \textit{og} might appear in very different contexts derived from words like \textit{dial og ues} and \textit{pseud og ene}.
We propose a system which prepares subwords for their downstream use by baking in the contextualized signal \textbf{at the vocabulary creation step}.
Our model, \sage, uses the \skipgram{} objective~\cite{mikolov2013distributed} over a corpus as the basis for iteratively eliminating candidate subwords from an initial large vocabulary
until the desired vocabulary size has been reached.
As \autoref{tab:examples} shows, \sage{} succeeds in removing the ambiguous \textit{og} token, facilitating distinct contextualization procedures for the example sentences (taken from Wikipedia).

We present our algorithm, \sage{}, which is predicated on iterative pruning of contextually noisy tokens from the vocabulary, and compare its effects on token properties and context cohesion with \bpe{} both in- and out-of-domain, in English and Turkish.
We then evaluate its performance on downstream tasks by training a BERT-based LLM~\cite{devlin-etal-2019-bert} on a vocabulary produced by both tokenizers in both languages, demonstrating substantial improvements on most English GLUE tasks and on NER, as well as Turkish NLI and NER.
We emphasize that as opposed to most current tokenizer variants, our model is a \say{plug and play} substitution for any subword token vocabulary, requiring no modification in the inference protocol (or code) when pre-training or applying an applicable LLM from a popular shared library.\footnote{Our code and models are available at \url{www.github.com/MeLeLbgu/SaGe}.}

\section{Subword Vocabulary Creation}
\label{sec:background}

The methods used to tokenize corpus in order to later assign tokens with continuous vectors, or \textit{embeddings}, have evolved over the years.
Initially, each word in the corpus was assigned its own embedding~\cite{collobert2008unified,mikolov2013distributed}.
OOVs, i.e.~words not appearing in the original training corpus or below a certain frequency threshold, would receive a special (but identical) \say{UNK} vector.
Subword tokenizers~\cite{schuster2012japanese,Wu-2016-wordpiece} were introduced to alleviate this issue, allowing segmentation of all text into embeddable units (assuming no unseen characters, a much more relaxed constraint for languages using alphabetical scripts).
The training process used to create a subword vocabulary from which the model then decodes text input involves optimizing an encoding objective over a large corpus.
To date, all tokenizers used in practice in large models focus on efficiency and information-theoretic objectives, and reduce the corpus to a unigram frequency count of space-delimited words, reducing calculation time but losing all contextual signal.
\sage{} reintroduces the contextual dependencies between words into vocabulary creation via a two-stage process, namely over-application of \bpe{} followed by iterative pruning using ideas inspired by \unigramlm{} and \skipgram{}.
We briefly present these algorithms before tying them together into \sage{}.

\begin{algorithm}[t]
\small
\caption{Byte-pair encoding vocabulary creation~\cite{Gage1994ANA,sennrich-etal-2016-neural}}
\label{alg:bpe}
\hspace*{\algorithmicindent} \textbf{Input:} Corpus $C$, Vocabulary final size $V$. \\
 \hspace*{\algorithmicindent} \textbf{Output:} Vocabulary $\mathcal{V}$ of size $V$ (ordered).
 \begin{algorithmic}[1]
 \Procedure{\bpe}{$C, V$}
    \State $\mathcal{V} \gets$ All unique characters in $C$
    \While{$|\mathcal{V}| < V$} \Comment{Merge tokens}
        \State $\langle t_{L}, t_{R} \rangle \gets$ Most frequent bigram in $C$
        \State $t_{NEW} \gets t_{L} \oplus t_{R}$ \Comment{Make new token}
        \State $\mathcal{V} \gets \mathcal{V} \oplus [t_{NEW}]$
        \State $C.$ReplaceAll$(\langle t_{L}, t_{R} \rangle, t_{NEW})$
    \EndWhile \\
    \Return{$\mathcal{V}$}
\EndProcedure
 \end{algorithmic}
\end{algorithm}

\paragraph{Byte-Pair Encoding.}
The \bpe{} algorithm creates a vocabulary \say{bottom-up}, starting with all single characters from the alphabet, iteratively adding tokens until reaching the desired vocabulary size.
In each iteration, the added token is the concatenation of the most frequent adjacent pair of existing tokens (see Algorithm~\ref{alg:bpe}).
The default setting of the algorithm's most popular implementation~\cite{kudo-richardson-2018-sentencepiece} restricts token addition within word boundaries, facilitating training from unigram frequencies.
In addition, LLM tokenizers using \bpe{}~\cite{liu2019roberta,radford2019language,wolf-etal-2020-transformers} decode sequences not by applying merges by order of the vocabulary, as originally dictated by the algorithm, but through greedy largest-subsequence left-to-right inference.
 
\paragraph{Unigram Language Model.}
\unigramlm{} offers a top-down vocabulary creation process, starting with an initial vocabulary of all substrings in the input corpus and pruning tokens iteratively until reaching the desired vocabulary size.
The pruning procedure involves calculating the overall unigram likelihood of the corpus with the current vocabulary versus a vocabulary lacking the candidate pruning token (see Algorithm~\ref{alg:unilm} for details), which we refer to as the \textbf{ablation objective}.
Under this system, decoding is ideally performed by considering probabilities of all possible segmentations using, e.g., the Viterbi algorithm; again, common practice is to use left-to-right greedy decoding.

\begin{algorithm}[t]
\small
\caption{\unigramlm{} vocabulary creation \cite{kudo-2018-subword}. $n \argmin_X$ denotes the $n$ bottom-ranked elements in $X$.}
\label{alg:unilm}
\hspace*{\algorithmicindent} \textbf{Input:} Corpus $C$, Vocabulary final size $V$, pruning batch size $k$. \\
 \hspace*{\algorithmicindent} \textbf{Output:} Vocabulary $\mathcal{V}$ of size $V$.
 \begin{algorithmic}[1]
 \Procedure{\unigramlm}{$C, V$}
    \State $\mathcal{V} \leftarrow$ All substrings occurring more than once in $C$
    \While{$|\mathcal{V}| > V$} \Comment{Prune tokens}
        \State $X^{(j)} \gets$ tokenize($C$, $\mathcal{V}$)
        \State $\mathcal{L}(\mathcal{V}) \gets \displaystyle \sum_{j=1}^{|C|} \log \left(P(X^{(j)})\right)$
        \ForAll{$t \in \mathcal{V}$}: \Comment{Calculate ablation objective}
        \State $loss_{t} \gets \mathcal{L}(\mathcal{V} \setminus \{t\}) - \mathcal{L}(\mathcal{V})$ 
        \EndFor
        \State $\mathcal{P} \gets \min(k, |\mathcal{V}|-V)\argmin_{t\in\mathcal{V}}(loss_t)$
        \State $\mathcal{V} \gets \mathcal{V} \setminus \mathcal{P}$ \Comment{Prune}
   \EndWhile \\
    \Return{$\mathcal{V}$}
\EndProcedure
 \end{algorithmic}
\end{algorithm}

\paragraph{Skipgram Objective.}
The \skipgram{} objective~\cite{mikolov2013distributed} formalizes the relation between a target token $t$ and its context, asking whether context tokens $c$ within a window $W_t$ of pre-defined size can be predicted from $t$.
These predictions are done via sigmoid activation over the inner product of embeddings trained for targets ($\vE^{(T)}$) and contexts ($\vE^{(C)}$). 
When aggregated over all tokens in a corpus, \skipgram{} can be used as a total likelihood measure, approximating its overall contextual cohesion:
\begin{equation}
    \small
    \label{eq:SG}
    \mathcal{L}(\mathcal{V},\mathcal{C}) = - \sum_{t \in \tok(\mathcal{C},\mathcal{V})} \sum_{c_j \in W_t} \log \left(\sigma(\vE^{(T)}_{t} \cdot \vE^{(C)}_{c_j})\right).
\end{equation}
As token vocabularies or their inference methods change, so do the target sequences and their contexts, resulting in differences in aggregated likelihood which can then act as scores comparing one tokenization to another.
We use this behavior as the ablation objective for \sage.

\section{\sage{} Vocabulary Creation}
\label{sec:model}

\begin{algorithm}[t]
\small
\caption{\sage{} vocabulary creation. $n \argmin_X$ denotes the $n$ bottom-ranked elements in $X$.}
\label{alg:sage}
\hspace*{\algorithmicindent} \textbf{Input:} Corpus $C$, Vocabulary final size $V$, basic tokenizer $\mathcal{T}$, overshoot factor $n$, pruning batch size $k$, likelihood recalculation frequency $m$, size of pruning candidate set $M$, embedding recalculation frequency $l$. \\
 \hspace*{\algorithmicindent} \textbf{Output:} Vocabulary $\mathcal{V}$ of size $V$.
 \begin{algorithmic}[1]
 \Procedure{\sage{}}{$C, V$}
    \State $\mathcal{V} \leftarrow \mathcal{T}(C, n \cdot V)$
    \State $i \leftarrow 0$ 
    \While{$|\mathcal{V}| > V$} 
        \If{$i \equiv 0\;(\bmod~{l\times m})$}
        \State $\vE^\mathcal{V} \gets $Word2Vec($\mathcal{V}$) \Comment{Embedding table}
        \EndIf
        \State $\mathcal{L}(\mathcal{V}, \mathcal{C}) \gets$ SGObj($\vE^\mathcal{V}, \mathcal{C}$) \Comment{Total likelihood (\ref{eq:SG})}
        \If{$i \equiv 0\;(\bmod~{m})$} \Comment{Update bottom set}
        \ForAll{$t \in \mathcal{V}$}:
        \State $loss_{t} \gets \mathcal{L}(\mathcal{V} \setminus \{t\}, \mathcal{C}) - \mathcal{L}(\mathcal{V}, \mathcal{C})$
        \EndFor
        \State $\mathcal{V}_{bot} \gets M\argmin_{t \in \mathcal{V}}(loss_t)$
        \Else \Comment{Update losses for bottom set}
        \ForAll{$t \in \mathcal{V}_{bot}$}:
        \State $loss_{t} \gets \mathcal{L}(\mathcal{V} \setminus \{t\}, \mathcal{C}) - \mathcal{L}(\mathcal{V}, \mathcal{C})$
        \EndFor
        \EndIf
        \State $\mathcal{P} \gets \min(k, |\mathcal{V}|-V)\argmin_{t\in\mathcal{V}_{bot}}(loss_t)$
        \State $\mathcal{V}_{bot} \gets \mathcal{V}_{bot} \setminus \mathcal{P}$ \Comment{Prune}
        \State $\mathcal{V} \gets \mathcal{V} \setminus \mathcal{P}$
        \State $i \gets i+1$
   \EndWhile \\
    \Return{$\mathcal{V}$}
\EndProcedure
 \end{algorithmic}
\end{algorithm}

\begin{table*}[t]
    \centering
    \small
    \begin{tabular}{lc} \toprule
        Sentence fragment & \dots{} use of an \textcolor{blue}{include} directive is when referring to \dots{} \\
        Tokenization using $\mathcal{V}$ & \_use \_of \_an \textcolor{brown}{\_includ} \textbf{[}\textcolor{brown}{e} \_direct \textbf{ive} \_is \_when\textbf{]} \_ref er r ing \_to \\
        Tokenization using $\mathcal{V}\setminus\{\texttt{\_includ}\}$ & \_use \_of \_an \textcolor{red}{\_inc l u} \textbf{[}\textcolor{red}{de} \_direct \textbf{ive} \_is \_when\textbf{]} \_ref er r ing \_to \\
        \bottomrule
    \end{tabular}
    \caption{The effect of retokenization on a context window of width 2 (in brackets) surrounding a target token (in bold). A left-side context token has been replaced as a result of an out-of-window vocabulary ablation.}
    \label{tab:retok}
\end{table*}

\sage{}\footnote{The name is not an acronym; it is intended to evoke SkipGram while maintaining the \say{suffix} of \emph{BPE}.} is a top-down tokenizer, following \unilm's general procedure, incorporating a \skipgram{} objective as its vocabulary trimming rule.
Given an initial vocabulary $\mathcal{V}$ and a corpus $\mathcal{C}$, \sage{} computes a \skipgram{} embedding space over $\mathcal{V}$ which provides it with an overall likelihood over $\mathcal{C}$ as in (\ref{eq:SG}).
It then proceeds to calculate the \emph{loss} of each token in the vocabulary were it to be removed, eliminating the tokens incurring minimal loss and re-tokenizing the corpus according to the updated vocabulary, repeating this procedure until reaching the desired vocabulary size $V$.
Having learned this vocabulary, downstream inference proceeds exactly as in the other segmentation-based methods, in a greedy left-to-right manner.
\sage{} can also be adapted to anticipate other decoding algorithms, by changing the re-tokenization steps accordingly.

In practice, applying the full process described above introduces multiple sources of considerable computational complexity:
for example, calculating the ablation objective for each token in each iteration produces a quadratic amount of calculations over the entire corpus;
recalculating embeddings for an updated vocabulary is similarly unreasonable to perform at each iteration.
We ameliorate these and other sources of complexity using a series of heuristics found in preliminary experiments to be minimally disruptive to precision of likelihood calculations.
We will now describe these heuristics, all depicted in Algorithm~\ref{alg:sage}.
First, instead of initializing the vocabulary as the full set of possible character sequences in the corpus, as in \unilm{}, we use any existing noncontextual tokenizer such as \bpe{} to learn a vocabulary larger than $V$ by a factor of $n$, and begin the pruning process from there.
Next, instead of removing a single token from the bottom of the loss-ranked vocabulary, we remove a batch of the $k$ bottom tokens each time, as does \unilm{}.\footnote{As in \unilm{} and other ablation-based vocabularies, single-character tokens are never removed from the vocabulary, in order to allow for all in-alphabet words to be tokenized.}
To avoid frequent loss recalculation, we recompute the entire likelihood set once every $m$ ablation steps, and only keep the bottom $M$ tokens as pruning candidates for the next $m$ steps.
Our preliminary experiments support this decision, as we found the ranked list of losses tends to stay relatively stable over dozens of batch-pruning iterations.
Lastly, to avoid the costly re-training of the embedding matrix for all tokens given the updated corpus, which only results in minor changes in likelihood during subsequent iterations, we only perform it every $l$ iteration batches, i.e.~after the ablation of $k\times{}m\times{}l$ tokens.
$n$, $k$, $l$, $m$ and $M$ are all algorithm hyperparameters tuned empirically based on desired runtime, corpus size and vocabulary size.

\paragraph{Contextual Loss.}
In order to calculate the per-token \skipgram{} likelihood loss, all sentences where a token $t$ occurs need to be re-segmented according to $\mathcal{V}\setminus\{t\}$, and their new likelihoods recorded.
To support performing this calculation on a large scale, we maintain a mapping of tokens to sentences containing them, as well as these sentences' current likelihoods.
This must be done at the sentence level rather than the window level, since a remaining suffix from an out-of-window re-tokenization may combine with in-window characters and form different token sequence replacements at a given stage.
Consider the example in \autoref{tab:retok}, where re-tokenization results in the replacement of a context token for a distant target. 

\paragraph{Negative Sampling.}
The original \skipgram{} objective uses negative samples to estimate context probabilities.
Since our application of \skipgram{} within the vocabulary creation algorithm (independent of the embeddings training procedure) includes only likelihood estimation with no parameter updates, we do not sample negative tokens, a process which would introduce substantial noise and complexity.

\section{\sage{} Vocabulary Properties}
\label{sec:properties}

For an analysis of our modified algorithm's advantages, we trained vocabularies of a pre-determined size using both \bpe{} and \sage.
We selected $|\mathcal{V}|=16,000$, and obtained corpora for English (750,000 lines from the August 2022 Wikipedia dump) and Turkish (the entire text of the September 2022 Wikipedia dump), opting for languages that share the Latin alphabet but differ in family (Indo-European vs. Turkic) and, crucially, in morphological properties: English is a low-exponence, low-synthesis language, while Turkish features multiple inflectional exponence and high verbal synthesis, as well as highly agglutinative morphology~\cite{wals-21,wals-22}.
We used the following hyperparameter settings to compute the vocabularies: Initial vocab size $20,000$ (or $n=1.25$), $l = 4$, $k = 100$, $M = 1500$, $m = 10$.
We used the Gensim package to train the \skipgram{} models~\cite{rehurek2011gensim}, and Sentencepiece~\cite{kudo-richardson-2018-sentencepiece} to obtain the initial \bpe{} vocabularies.\footnote{Since \bpe{} augments its vocabulary iteratively, the baseline \bpe{} vocabulary is a proper subset of that used to initialize \sage.}
More hyperparameters are detailed in \autoref{app:hyppar}.

We present an analysis of the resulting vocabularies, highlighting the advantages and trade-offs exhibited by context-based subword tokenization.
Generally speaking, most of the tokens discarded from \sage's initial vocabulary appear in the baseline \bpe's final vocabulary.
Among the differences between the vocabularies are many short tokens that appear in \bpe{}'s but not \sage's, proper substrings of longer tokens also appearing in the \bpe{} vocabulary.
This is due to \bpe's bottom-up merge table construction, which forces retention of the entire chain of tokens created: if \emph{the} is part of the vocabulary, either \emph{th} or \emph{he} must also be there.
While essential for the original intended decoding process, actual implementations of greedy decoding have no need for this property.
\sage's initial vocabulary shares this characteristic, but the trimming process allows any token to be ablated, including those in the middle of merge chains.
Another difference found between the vocabularies is the strong preference of \sage{} for word-initial tokens.
83\% of the tokens that appear in \sage{}'s vocabulary but not in \bpe's are word-initial, compared to only 22\% of the \bpe{}-only tokens.
This is reasonable, since a token surviving \sage{}'s ablation steps exhibits high loss for the condition of its removal, which is arguably the case when a nearby target word needs to predict a word-initial context.

\begin{figure}
    \centering
    \includegraphics[width=\columnwidth]{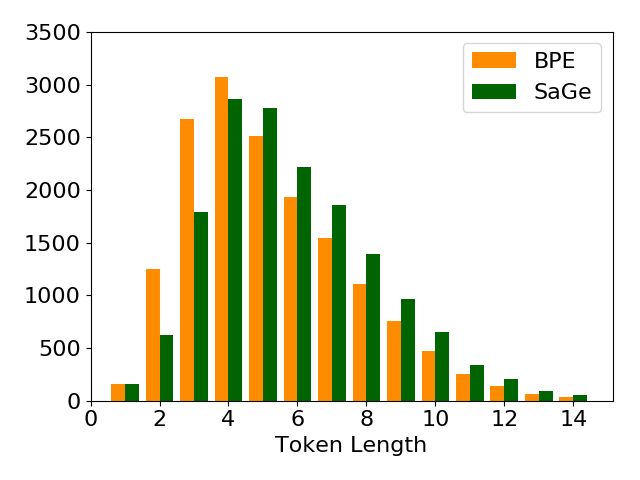}
    \caption{Token length distribution of \bpe{}'s vocabulary vs. \sage{}'s on English.}
    \label{fig:lengths}
\end{figure}

\begin{table}
    \centering
    \begin{tabular}{llp{1.5cm}rr}
        \toprule
        \multicolumn{5}{c}{More frequent in} \\
        \multicolumn{2}{c}{\sage{}} & & \multicolumn{2}{c}{\bpe{}} \\
        \midrule
        e & s & & es & ic \\
        ing & ist & & ings & ff \\
        ation & ate & & ations & ates \\
        \bottomrule
    \end{tabular}
    \caption{Tokens with high difference in frequency between tokenizations (English models).}
    \label{tab:freqs}
\end{table}

\paragraph{Token Length.}
\autoref{fig:lengths} shows a histogram of token lengths (in characters) for the 16,000-token \sage{} and \bpe{} vocabularies in English (results on Turkish are similar).
\sage{} clearly selects longer tokens for its vocabulary, again a sensible outcome given their higher chance of being contextually coherent.
The difference is most stark with tokens of length 2 and 3;
when considering only tokens appearing in exactly one of the final vocabularies, we find that 56\% of \bpe{}-only tokens are of length 2 and 3, while 55\% of \sage{}-only tokens are of length 5 and above.

\paragraph{Token Frequency.}
We compute the frequency of tokens in the encoding form of the English training corpus, once using \sage{} vocabulary and once using \bpe's.
In \autoref{tab:freqs} we show some of the tokens with the biggest difference in frequency between \sage{} and \bpe{} tokenizations.
We can see \sage{} reverts to single-character tokens considerably more often than \bpe{} (also demonstrated in the last example in \autoref{tab:examples}).
We view this as a feature of context-based tokenization---its vocabulary is partitioned between (mostly short) tokens that are highly ambiguous in context and (mostly long) tokens that have coherent contexts.
At the same time, \bpe{} is rife with tokens that are medially ambiguous contextually, whose resulting embeddings can be neither useful nor completely ignorable, adding noise to the representation sequences.
As a result, \sage{} breaks down complex suffixes, which in English are compositional, into their constituent morphology.
The suffix \emph{ings} is thus dismantled to \emph{ing s}, whereas \bpe{} reserves a token for it, mostly unhelpful in itself.

\begin{figure}
    \centering
    \includegraphics[width=\columnwidth]{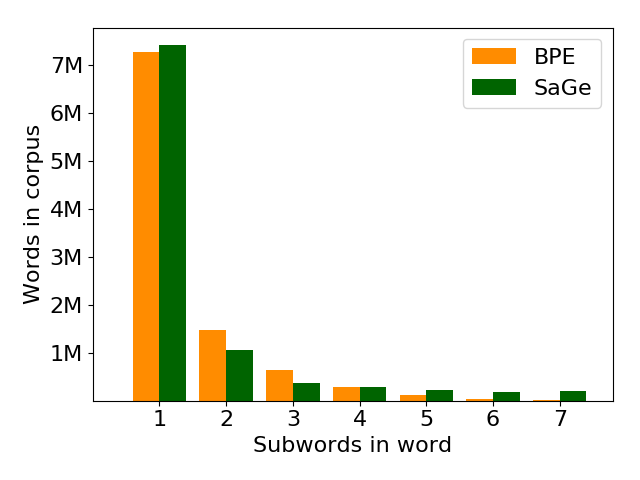}
    \caption{Number of subwords required to tokenize a word, collected over the original English training corpus.}
    \label{fig:fertility}
\end{figure}

\begin{figure*}
    \centering
    \includegraphics[width=0.7\textwidth]{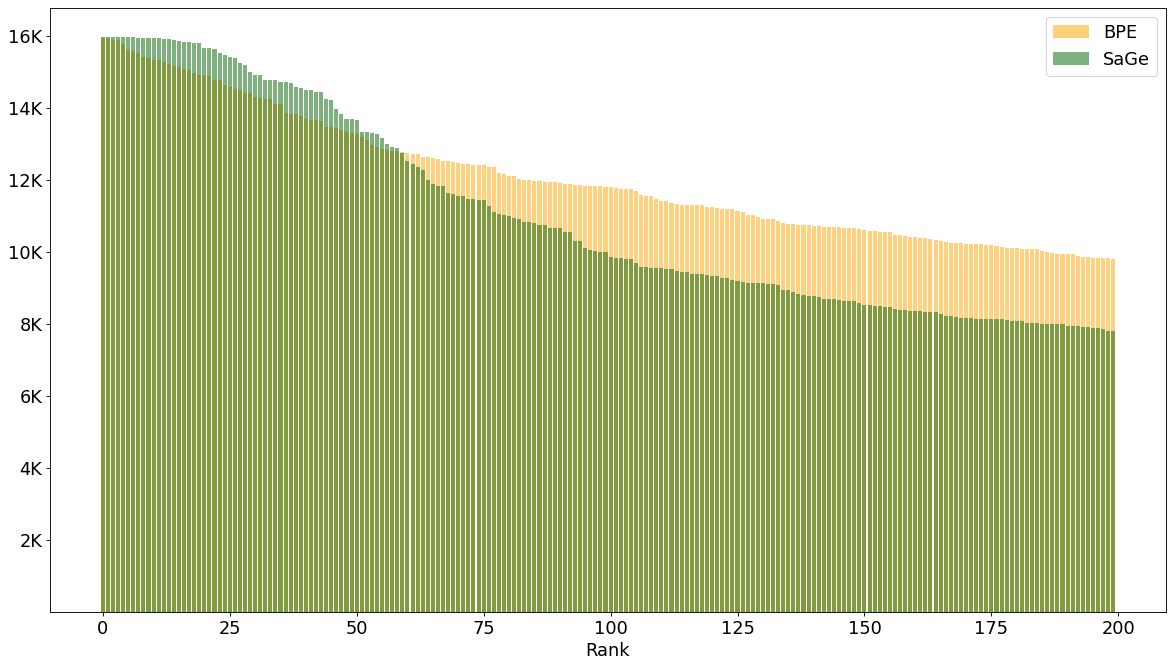}
    \caption{Number of distinct neighbors each token encounters in a width-5 window, top 200, Turkish.}
    \label{fig:exponence}
\end{figure*}

\paragraph{Subword Fertility.}
Fertility, as defined in the statistical machine translation literature, refers to the average number of subwords produced per tokenized word.
\autoref{fig:fertility} exhibits a histogram of all English corpus words by their subword length, using the \bpe{} vocabulary and the \sage{} vocabulary.
Although \sage{} retains more words as single tokens, it trades them off with more words having five subwords or more, compared with \bpe's abundance of words with 2 and 3 subwords.
This follows the trend described so far, of \sage{}'s preference for dismantling unknown words into meaningless single-character tokens rather than confusing, ambiguous length-2 and length-3 tokens.
We believe that \bpe's behavior harms text understanding in suggesting that these ambiguous fragments (consider \say{og}) have some meaning that an LLM can try and learn, whereas \sage's single-character breakdown indicates a word that's truly unknown and cannot be inferred by composing constituent in-vocab subwords.

Fertility translates to a trade-off in \textbf{encoding efficiency} to \sage's contextual advantage: a sample of 150K lines from English Wikipedia is encoded by 4 million \bpe{} tokens, optimizing only an information-theoretic objective, whereas \sage{} produces 4.5 million.
Having said that, this inefficiency might be further offset during LLM pre-training: we propose that contextually coherent tokens will require fewer update steps in order to achieve useful embedding parameters, helping the model converge faster compared to \bpe{} tokens.
We leave testing this hypothesis to future work.

\begin{figure}
    \centering
    \includegraphics[width=0.9\columnwidth]{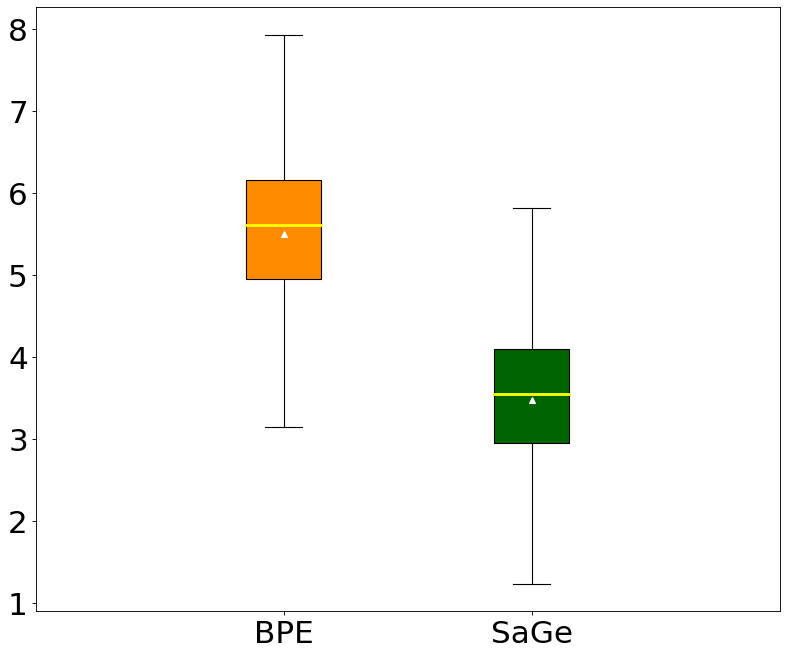}
    \includegraphics[width=0.9\columnwidth]{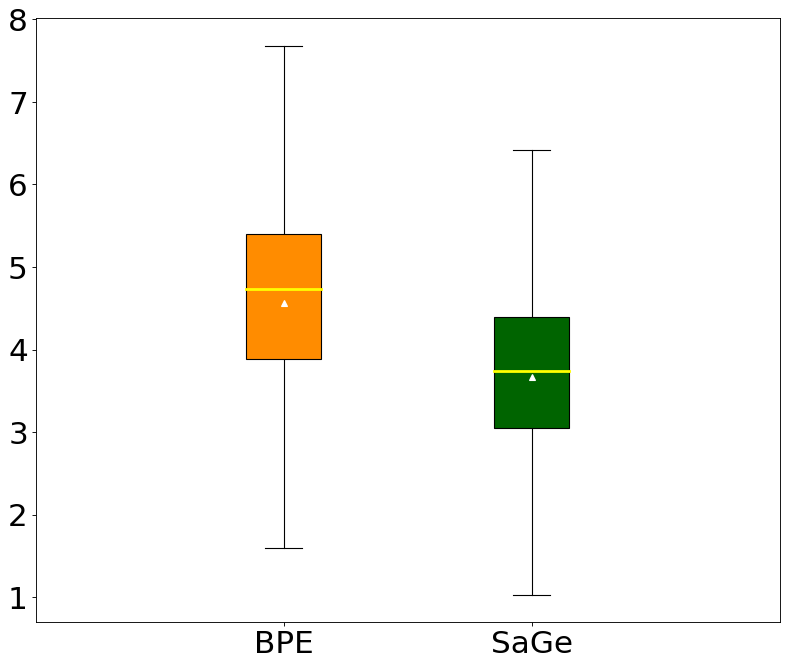}
    \caption{Distribution of token neighbors/frequency ratio for a width-5 window in English (top) and Turkish (bottom); \bpe{} (left) and \sage{} (right)}
    \label{fig:freq-cont}
\end{figure}

\paragraph{Contextual Exponence.}
To determine the degree to which \sage{} effectively optimizes tokens' contextual soundness, which is its ultimate goal, we plot the number of distinct neighbors each token encounters throughout the training corpus, ranked from high to low, in \autoref{fig:exponence}.
The very top of the ranking is occupied by single-character tokens which are context-null by design, which \sage{} makes the most of by placing in almost all contexts.
After a few dozen tokens, \sage{}'s context counts dip below \bpe{}'s, a trend which continues all the way through the vocabulary, making up a more contextually coherent set.
These findings hold for English as well as Turkish, and replicate when taking a context window of size 2, different from that used during \sage{} construction.

These findings can arguably be attributed to a frequency artifact, where \sage{} simply outputs more tokens with lower frequency in order to provide them with fewer contexts.
We thus present a normalized analysis in \autoref{fig:freq-cont}, depicting the ratio between each token's number of unique neighbors and its frequency, distributed over the entire vocabulary.
\sage{} provides substantially lower ratios in both languages, supporting our original claim.

\subsection{Robustness to Domain Change}

\begin{figure}
    \centering
    \includegraphics[width=0.9\columnwidth]{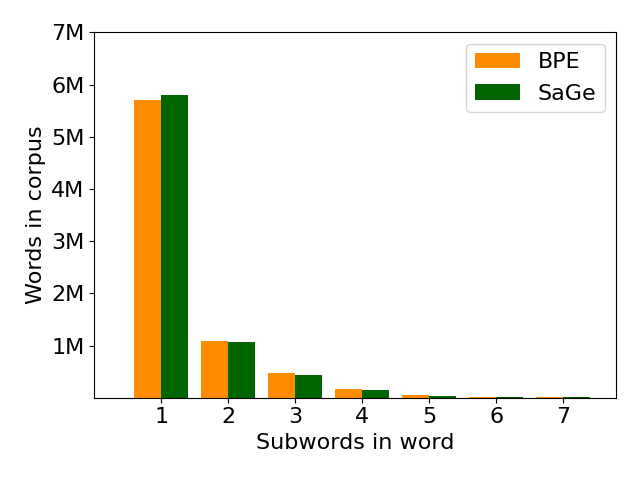}
    \caption{Number of subwords required to tokenize a word using the original Wikipedia-trained vocabularies, collected over a English Quora questions corpus.}
    \label{fig:fertility-domain-transfer}
\end{figure}

\begin{figure}
    \centering
    \includegraphics[width=0.8\columnwidth]{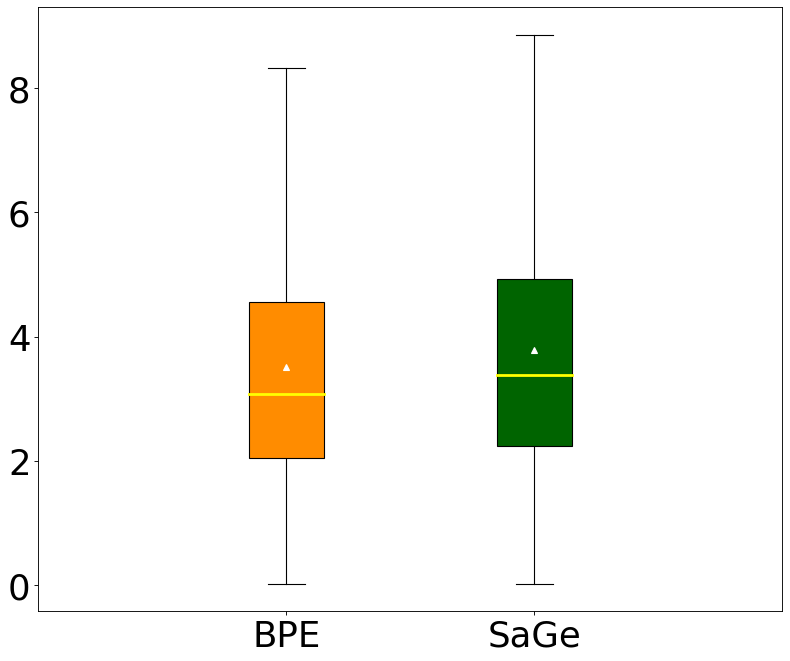}
    \caption{Distribution of token neighbors/frequency ratio for a width-5 window in English, based on a Wikipedia-trained vocabulary and collected over a English Quora questions corpus.}
    \label{fig:freq-cont-domain-transfer}
\end{figure}

One possible limitation of the \sage{} objective is that it increases the reliance on the original training corpus compared to word-count-only algorithms.
In and of itself, this should not necessarily be viewed as a problem, assuming the collected corpus is a faithful representative of an LLM's use case.\footnote{Indeed, existing literature recommends adding pre-training steps on new domains before fine-tuning models for them~\cite[e.g.,][]{han-eisenstein-2019-unsupervised}.}
To this end, we collected comparable corpora from non-Wikipedia domains and ran our analysis on the \sage{} and \bpe{} vocabularies trained on Wikipedia.
Our findings suggest that while \sage{} loses its relative advantage in context-dependence over \bpe{}, it does not fall behind it (i.e.~it has not overfit to the Wikipedia domain).
We present a fertility chart for an English corpus of 7.5M words from Quora questions\footnote{\url{https://huggingface.co/datasets/chenghao/quora_questions}} in \autoref{fig:fertility-domain-transfer}, depicting similar trends to that on Wikipedia (\autoref{fig:fertility}) but with smaller differences between \sage{} and \bpe{}; the neighbor-to-frequency ratio aggregation chart in \autoref{fig:freq-cont-domain-transfer} differs from \autoref{fig:freq-cont} (top) substantially but shows that \sage{} and \bpe{} tokens do not diverge significantly on this measure.
We repeated the experiment on English legal text centered on US congress bills~\cite{pileoflaw-hendersonkrass2022} and on a 2.6M-word Turkish corpus of online reviews,\footnote{\url{https://huggingface.co/datasets/cansen88/turkishReviews\_5\_topic}} and observed similar trends.

These results indicate that while a considerable amount of the longer tokens preferred by \sage{} were selected to optimize contextuality in the source domain, as it was designed to do, there is no \say{short blanket} effect for text originating in different domains.
This could either be due to wide-scope advantages of some of the tokens selected by \sage{}, or due to an intrinsic deficiency in \bpe's long-tail tokens, or a combination of both.

\begin{table*}
    \centering
    \begin{tabular}{lrrrrrrrp{0.2cm}r}
        \toprule
         & MRPC & MNLI & COLA & QNLI & SST2 & STSB & QQP & & XNLI$_{tur}$ \\
         & (F1) & (Acc \%) & (Matt.) & (Acc \%) & (Acc \%) & (Pear.) & (Acc \%) & & (Acc \%) \\
        \midrule        
       \bpe & .7918 & 62.76 & .0777 & 66.17 & 80.54 & .3094 & 82.75 & & 41.20 \\
        \sage & .8004 & 64.00 & .0985 & 74.83 & 79.85 & .3387 & 84.69 & & 46.46 \\
        \bottomrule
    \end{tabular}
    \caption{Performance on sequence-level tasks for BERT models trained on different 16k-size vocabularies. XNLI$_{tur}$ is Turkish, the rest are English GLUE tasks.
    All results averaged over three runs on the dev set with different seeds.}
    \label{tab:results}
\end{table*}

\section{Downstream Evaluation}
\label{sec:eval}

In order to evaluate the utility of our tokenization algorithm for major NLP tasks, we compare \sage{} to a \bpe{} vocabulary of the same size by means of pre-training a BERT-parameterized model~\cite{devlin-etal-2019-bert} using an expedited training scheme~\cite{izsak-etal-2021-train}.
We then evaluate the LLM's performance both on sequence classification via the English GLUE benchmark~\cite{wang-etal-2018-glue} and the Turkish partition of XNLI~\cite{conneau-etal-2018-xnli}, and on named entity recognition in English~\cite{wang-etal-2019-crossweigh} and Turkish~\cite{al2015polyglot}.
We use the default settings from Huggingface's library implementations of the fine-tuning processes~\cite{wolf-etal-2020-transformers} and do not perform hyperparameter tuning for either model.

We present our results on sequence-level tasks in \autoref{tab:results}.
\sage{} tokenization improves performance on nearly all tasks 
with particularly substantial improvements (1.3--8 accuracy points) on NLI datasets.
Results on NER are presented in \autoref{tab:results-NER}, again showing \sage{}'s dominance over \bpe.
Due to the length of the training pipeline leading from vocabulary creation through pre-training to fine-tuning, it is difficult to find individual examples where difference in tokenization leads to direct changes in prediction; we attribute the consistent overall gains in downstream performance mostly to the LLM pre-training step, where the design of \sage's context-friendly vocabulary enables a more coherent contextual signal to flow through the transformer layers during backpropagation.
We note that in general, our models fare worse on GLUE tasks compared to \newcite{izsak-etal-2021-train}.
We attribute this in part to the smaller token vocabulary size, and more substantially to the smaller pre-training corpus we used in our experiments.

\begin{table}
    \centering
    \begin{tabular}{lrr}
        \toprule
         & English & Turkish \\
        \midrule
        \bpe & .7142 & .4660 \\ 
        \sage & .7502 & .5475 \\ 
        \bottomrule
    \end{tabular}
    \caption{Performance (F1) on NER tasks of BERT Turkish and English models trained on different subword vocabularies of size 16,000.
    All results averaged over three runs on the dev set with different seeds.
    }
    \label{tab:results-NER}
\end{table}

\section{Related Work}
\label{sec:related-work}

In recent years, a growing body of research has demonstrated the shortcomings of existing tokenization algorithms in the context of representing linguistic phenomena in different languages across different tasks~\cite{banerjee2018meaningless,klein-tsarfaty-2020-getting,hakimi-parizi-cook-2020-evaluating,rust-etal-2021-good,maronikolakis-etal-2021-wine-v,mielke2021between,hofmann-etal-2021-superbizarre}, as well as the statistical properties affecting their downstream performance~\cite{bostrom-durrett-2020-byte}.
Our work addresses the concerns raised in this line of work by introducing an improved subword vocabulary creation method which leverages the contextual aspects of the main intended use case, namely LLMs.
Previous work towards this goal includes algorithms which offer robustness within an existing subword vocabulary~\cite{provilkov-etal-2020-bpe,he-etal-2020-dynamic,hiraoka2022maxmatch}, necessitating modification of either training, inference, or both procedures in the context of LLMs.
Others have considered tuning the \emph{size} of a subword vocabulary~\cite{salesky2020optimizing}, or selecting from an enlarged set of possible segmentations~\cite{asgari2020subword}, for optimizing performance on downstream tasks.

Some alternative tokenization methods focus on the application of a model which considers the expected downstream tasks together with the pre-training corpus~\cite{hiraoka-etal-2020-optimizing}, to the degree of jointly optimizing the tokenizer with the downstream model~\cite{hiraoka-etal-2021-joint}. 
In addition to the massive changes in training and inference procedures this approach incurs, we note that it is difficult to apply to large contextualized models due to the long path from tokenization to prediction; \sage{} overcomes this problem by \say{nudging} only the LLM vocabulary itself towards a contextualization-friendly segmentation.

The concept of subword tokenization made its rise alongside that of contextualized representations, meaning that little work exists where \skipgram{} or other static models are trained over proper subword segmentations.
Recently, \newcite{kaushal2022tokens} did so for a proof-of-concept of a spelling prediction model, in lieu of training full LLMs. 
To our knowledge, no work to date has used a static embedding-based objective to score token sequence likelihood for a separate task (as we do for vocabulary trimming).

Finally, we acknowledge the recent efforts to do away with tokenization altogether, be it through character-only~\cite{clark-etal-2022-canine} or byte-only~\cite{xue-etal-2022-byt5} models, or through encoding characters visually and passing them through a vision model~\cite{salesky-etal-2021-robust,rust2022language}.
These represent an even more radical departure from the established application of LLMs, and we look forward to testing their abilities against our improved contextual subword tokenization methods.
We note that while these models have been facing issues regarding scaling, mostly on the decoding side, \sage{} vocabularies are ready to be used immediately within existing popular LLM implementations.
Furthermore, recent work has shown the limited utility of character-level transformers in semantic tasks, even for morphologically rich languages with nontrivial orthography-morphology relations~\cite{keren2022breaking}.

\section{Conclusion}

In this work, we introduced \sage, a context-aware tokenizer built using insights from \bpe, \unigramlm, and \skipgram, and showed that it achieves better results when used in an LLM-pre-train-then-fine-tune schema on two typologically distant languages on both the sequence and token levels.
We believe that further investigation into incorporating context in tokenization models can improve results even further, and intend to also extend our efforts toward other languages and writing systems, as well as to multilingual tokenizers.
For example, we plan to apply \sage{} in the context of Abjads like Hebrew and Arabic, as well as languages written in alphasyllabaries such as Devanagari.

Within \sage{} itself, there is room for improvement.
The algorithm is still relatively slow, taking roughly a day to run on a strong CPU, making it difficult to apply to a truly large corpus, to start from a larger initial vocabulary, or to conduct exhaustive search over the hyperparameters.
We intend to keep optimizing it, and continue evaluation against other subword and character-only schemas.

\section*{Limitations}

We acknowledge several limitations of \sage, a novel algorithm still in its development stages.
First, scaling the vocabulary creation framework up from corpus-level unigram statistics to context dependence incurs many points where linear factors turn into quadratic, and worse.
We introduced several heuristics to alleviate this issue in \S\ref{sec:model}, however \sage{} still takes longer to train compared to \bpe{} and other tokenizers, by roughly a factor of ten.
While having no effect on downstream pre-training and fine-tuning steps, it does mean hyperparameters are more difficult to tune.
Second, the prohibitive resources required to implement a full LLM pipeline has limited our downstream evaluation setup to ten individual tasks on two languages.
Ideally, as more languages with more diverse scripts and typological properties are examined, better generalizations can be made about the utility of integrating context into subword tokenizer vocabularies.
Finally, we still do not have a well-formed theory of integrating multiple domains, languages, or scripts together into a single vocabulary.
This question has interested researchers in recent years~\cite[e.g., ][]{chung-etal-2020-improving,rust-etal-2021-good,zhang-etal-2022-robust}, yet a tokenizer-internal solution (as opposed to data balance manipulation) still seems to have eluded the community.
This question affects \sage{} more than other tokenizers, given its reliance on context, which changes starkly when considering multiple sources of text in unison.

\section*{Acknowledgments}
We thank Jacob Eisenstein and Cassandra Jacobs for work on earlier versions of the high-level idea, and Timo Schick and L{\"u}tfi Kerem Senel for fruitful conversations in early stages of the project.
We thank Marco Cognetta, Michael Elhadad, Omer Levy, and attendees of ISCOL 2022 for comments and suggestions on more recent versions of the work.
We thank the reviewers for their helpful comments.
We thank Kaj Bostrom, Peter Izsak, and Tamar Levy for helping us obtain and operate resources for training our models.

\bibliography{anthology,sage}
\bibliographystyle{acl_natbib}

\clearpage

\appendix

\section{Hyperparameters}
\label{app:hyppar}

In \autoref{tab:hyperparameters_vocab}, \ref{tab:hyperparameters_bert}, and \ref{tab:hyperparameters_finetuning}, we present the hyperparameters used for training the various elements in our experiments.

\begin{table}[t]
\centering
 \begin{tabular}{lr} 
 \midrule
 Final Vocab Size & 16K\\
 Initial Vocab Size & 20K\\
 $k$ $($tokens to prune each batch$)$ & 100 \\
 $M$ $($size of pruning candidate set$)$ & 1500 \\
 $m$ $($likelihood recalculation frequency$)$ & 10 \\
 $l$ $($embedding recalculation frequency$)$ & 4 \\
 $\sage$ window size & 5 \\
 Word2Vec window size & 5 \\
 Word2Vec vector dimension & 50 \\
 Word2Vec negative samples & 15 \\ 
 \bottomrule
 \end{tabular}
\caption{Hyperparameters for vocabulary creation.}
\label{tab:hyperparameters_vocab}
\end{table}

\begin{table}[t]
\centering
 \begin{tabular}{lr} 
 \midrule
 layer\_norm\_type & pytorch \\
 model\_type & bert-mlm \\ 
 hidden\_act & gelu \\
 hidden\_size & 1024 \\
 num\_hidden\_layers & 24 \\
 num\_attention\_heads & 16 \\
 intermediate\_size & 4096 \\
 hidden\_dropout\_prob & 0.1 \\
 attention\_probs\_dropout\_prob & 0.1  \\
 encoder\_ln\_mode & pre-ln \\
 lr & 1e-3 \\
 train\_batch\_size & 4032 \\
 train\_micro\_batch\_size\_per\_gpu & 32 \\
 lr\_schedule & time \\
 curve & linear \\
 warmup\_proportion & 0.06 \\
 gradient\_clipping & 0.0 \\
 optimizer\_type & adamw \\
 weight\_decay &  0.01 \\
 adam\_beta1 & 0.9 \\
 adam\_beta2 & 0.98 \\
 adam\_eps & 1e-6 \\
 total\_training\_time & 24.0 \\
 optimizer\_type & adamw \\
 validation\_epochs & 3 \\
 validation\_epochs\_begin & 1 \\
 validation\_epochs\_end & 1 \\
 validation\_begin\_proportion & 0.05 \\
 validation\_end\_proportion & 0.01 \\
 validation\_micro\_batch & 16 \\
 deepspeed & yes \\
 data\_loader\_type & dist \\ 
 \bottomrule
 \end{tabular}
\caption{Hyperparameters for pre-training BERT-architecture models using the academic-budget-bert code~\cite{izsak-etal-2021-train}.}
\label{tab:hyperparameters_bert}
\end{table}

\begin{table}[t]
\centering
 \begin{tabular}{lr} 
 \midrule
 max\_seq\_length & 128 \\
 evaluation\_strategy & steps \\ 
 per\_device\_train\_batch\_size & 16 \\
 gradient\_accumulation\_steps & 1 \\
 per\_device\_eval\_batch\_size & 16 \\
 learning\_rate & 5e-5 \\
 weight\_decay & 0.1 \\
 max\_grad\_norm & 1.0 \\
 lr\_scheduler\_type & polynomial  \\
 warmup\_steps & 50 \\ 
 \bottomrule
 \end{tabular}
\caption{Hyperparameters for fine-tuning tasks using scripts from the academic-budget-bert package.}
\label{tab:hyperparameters_finetuning}
\end{table}

\section{Computing Resources}
\label{app:compute}

For our experiments we used Quadro RTX 8000 GPU.

\end{document}